\documentclass{article}
\usepackage{ijcai21}

\usepackage{titlesec}

\usepackage{times}

\usepackage{soul}
\usepackage{url}
\usepackage[hidelinks]{hyperref}
\usepackage[utf8]{inputenc}
\usepackage[small]{caption}
\usepackage{graphicx}
\usepackage{amsmath}
\usepackage{booktabs}
\usepackage{amsfonts}
\usepackage{bm}
\usepackage{enumitem}
\usepackage{etoolbox}
\usepackage{makecell}
\usepackage{microtype}
\usepackage{multirow}
\usepackage{tabularx}

\newcommand{\tabref}[1]{Tab.~\ref{tab:#1}}
\newcommand{\figref}[1]{Fig.~\ref{fig:#1}}
\newcommand{\secref}[1]{Sec.~\ref{sec:#1}}

\titleformat{\subsubsection}[runin]{\normalsize\bf\raggedright}{}{}{}[]
\titlespacing{\subsubsection}{0pt}{1pt plus 0.5pt minus 0.5pt}{0.5em plus 0.25em minus 0.25em}

\apptocmd{\thebibliography}{\setlength{\itemsep}{0.5pt}}{}{}

\title{Tool- and Domain-Agnostic Parameterization of Style Transfer Effects \\ Leveraging Pretrained Perceptual Metrics}

\author{
Hiromu Yakura$^{1,2}$\footnote{Contact Author}\and
Yuki Koyama$^2$\And
Masataka Goto$^2$\\
\affiliations
$^1$University of Tsukuba, Japan\\
$^2$National Institute of Advanced Industrial Science and Technology (AIST), Japan\\
\emails
\{hiromu.yakura, koyama.y, m.goto\}@aist.go.jp
}

\makeatletter
\apptocmd\@maketitle{{\herofigure{}\par}}{}{}
\makeatother
\begin{document}

\newcommand\herofigure{
  \centering
  \vspace{-10pt}
  \includegraphics[width=0.9\textwidth]{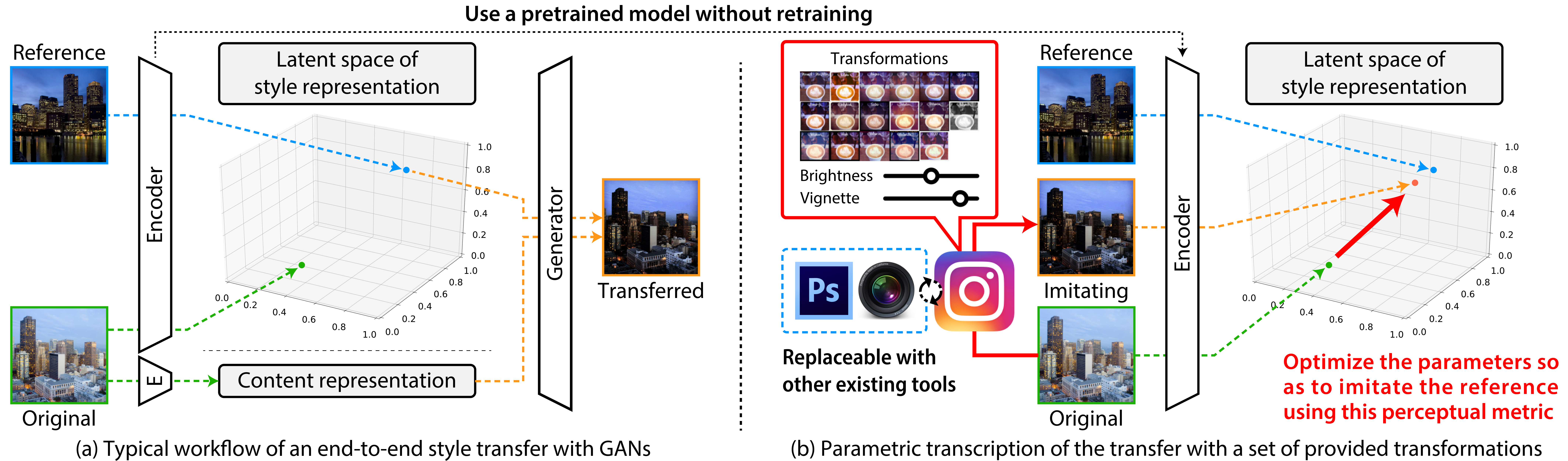}
  \vspace{3pt}
  \captionof{figure}{
    (a) Deep style transfer produces high-fidelity results but would be hard to utilize for exploratory design as it is performed only in an end-to-end manner.
    (b) Our framework transcribes the style transfer effect into a set of parametric transformations available in a tool the user is familiar with (e.g., Instagram), by which it encourages further exploration.
  }
  \vspace{14pt}
  \label{fig:teaser}
}

\maketitle

\begin{abstract}
Current deep learning techniques for style transfer would not be optimal for design support since their ``one-shot'' transfer does not fit exploratory design processes.
To overcome this gap, we propose \emph{parametric transcription}, which transcribes an end-to-end style transfer effect into parameter values of specific transformations available in an existing content editing tool.
With this approach, users can imitate the style of a reference sample in the tool that they are familiar with and thus can easily continue further exploration by manipulating the parameters.
To enable this, we introduce a framework that utilizes an existing pretrained model for style transfer to calculate a perceptual style distance to the reference sample and uses black-box optimization to find the parameters that minimize this distance.
Our experiments with various third-party tools, such as Instagram and Blender, show that our framework can effectively leverage deep learning techniques for computational design support.
\end{abstract}

\section{Introduction}
\label{sec:introduction}

Deep learning has enabled high-level manipulation of various media contents.
In particular, generative adversarial networks (GANs) \cite{DBLP:conf/nips/GoodfellowPMXWOCB14} paved the way not only for generating contents but also for modifying them via style transfer \cite{8732370}.
However, such end-to-end style transfer techniques would not be optimal for supporting users' design processes since they do not offer sufficient explorability.
More specifically, our design processes are often exploratory, that is, each is an open-ended journey starting with an under-specified goal \cite{DBLP:journals/tog/TaltonGYHK09}.
Through iterative and nonlinear exploration, we refine our understanding of the space of possible designs and establish the precise form of the final product in an opportunistic and serendipitous manner \cite{DBLP:conf/bcshci/Tweedie95,DBLP:conf/uist/TerryM02}.
Exploratory design is often compared to brainstorming and ideation because it would increase the potential for innovation and creativity \cite{DBLP:journals/aim/Gero90}.

For example, let us consider providing a current photo style transfer technique as a photo editing tool.
The tool can make a photo a user has taken (hereinafter referred to as an \textit{original} photo) look like a particular professional shot (hereinafter referred to as a \textit{reference} photo).
However, it cannot handle more flexible requests that occur in the nonlinear exploratory processes, such as making the contrast of the original photo similar to that of the reference photo but keeping the brightness unchanged, or exploring further variations by manually changing the strength of the transfer effect.

\begin{figure}[t]
    \centering
    \includegraphics[width=0.8\columnwidth]{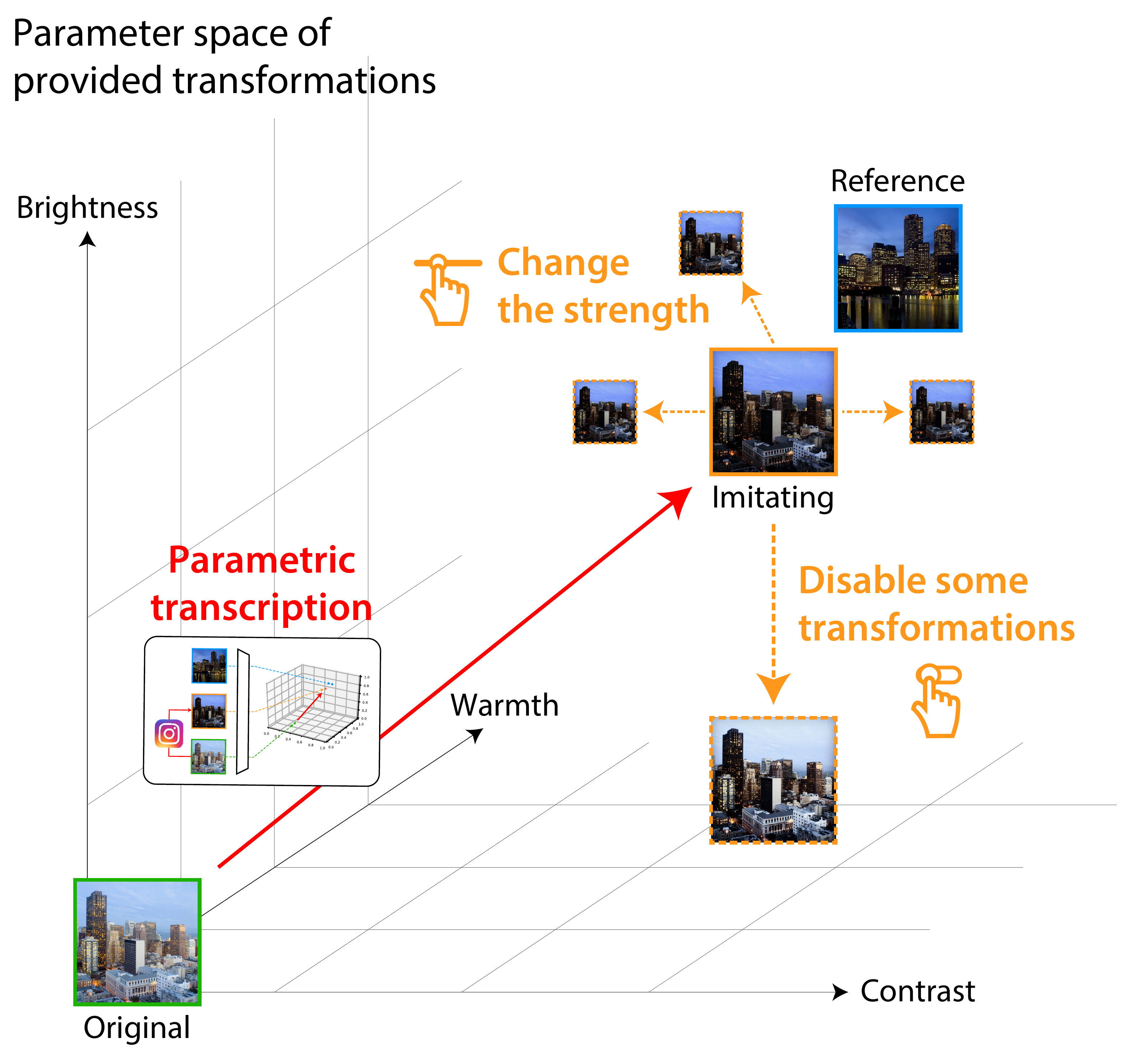}
    \caption{Since our framework shows the parameters of provided transformations to imitate style transfer, the user can further explore the variations by manipulating the parameters in the parameter space (e.g., disabling some transformations or changing the strength).}
    \label{fig:editable}
\end{figure}

To overcome this situation, we propose a novel approach, \emph{parametric transcription} of an end-to-end style transfer effect, which aims to provide a user with the best parameters of given parametric transformations to imitate the effect.
It is designed based on three key requirements we have identified to leverage deep learning techniques for exploratory design processes: serving parametric, transparent, and non-destructive transformations.
To satisfy them, it adopts parametric transformations from an existing editing tool that the user is familiar with, enabling the user to easily and intuitively explore the variations of the style-imitating result after transcription.
In photo editing, for example, instead of directly generating a ``style-baked'' image by an end-to-end style transfer, we show the user how to set parameters (such as brightness and contrast) and which photo filters to use to best imitate the style within Instagram.
Then, the user can further explore the variations by manipulating the parameters, as depicted in \figref{editable}.
Importantly, various existing tools can be used as the source of transformations to be used, and thus editable attributes are not limited a priori.

This paper introduces a universal framework for achieving parametric transcription (\figref{teaser}), which is tool- and domain-agnostic.
Its key components are a \emph{perceptual metric} retrieved from an existing pretrained model and \emph{black-box optimization}.
For the perceptual metric, we use the latent representation of content-invariant styles already acquired by the pretrained style transfer model.
By minimizing the distance between the latent representations of the transformed result and the reference sample, our framework finds the optimal parameters of the provided transformations for imitating the style transfer effect.
This optimization problem is not likely to be differentiable, but the adoption of black-box optimization \cite{DBLP:conf/kdd/AkibaSYOK19} makes it tractable.
Furthermore, black-box optimization enables users to employ various existing tools as sources of transformations.

To demonstrate the applicability of our framework, we conducted experiments in two different scenarios using popular smartphone apps: transcribing photo style transfer into Instagram and facial makeup transfer into SNOW.
In both scenarios, our framework was confirmed through subjective evaluations to be able to produce feasible results whose quality is comparable to those produced by humans.
In addition, to show that our framework is easily adapted to scenarios other than transcribing style transfers as well, we conducted an experiment in a pseudo-generative scenario performing text-to-object composition.
In this scenario, the original input is not given by a user but automatically selected from a set of candidates, which is done by parameterizing the selection and integrating it into the optimization process.
Based on these experiments, we discussed how parametric transcription can pave the way for creativity support tools \cite{DBLP:journals/cacm/Shneiderman07a} that benefit from the recent advent of deep learning techniques.

Our aim is not to present a new deep learning model for style transfer but to present a new general way to benefit from various existing pretrained models without additional training or data.
Our results with third-party tools validate the tool- and domain-agnostic applicability of our framework, suggesting that we can also incorporate it with models to appear in the future.
We believe that this paper shows a new paradigm of human-AI collaboration, especially in exploiting end-to-end deep learning techniques in a human-centric manner.

\section{Background}
\label{sec:background}

Cross-domain style transfer typically involves two networks: an encoder to map an input from one domain onto its latent representation and a generator to map the latent representation onto an output styled after another domain \cite{DBLP:conf/nips/LiuBK17}.
Formally, assuming that $\mathcal{X}$ and $\mathcal{Y}$ are two domains with different styles and $\mathcal{Z}$ is a latent space, an encoder $E_{\mathcal{X}}\!\!: \! \mathcal{X} \! \rightarrow \! \mathcal{Z}$ and a generator $G_{\mathcal{X}}\!\!: \! \mathcal{Z} \! \rightarrow \! \mathcal{X}$ are trained so that the transferred result of an original input $\bm{x} \! \in \! \mathcal{X}$ is obtained as $\bm{y} = G_{\mathcal{Y}} \! \left( E_{\mathcal{X}} \! \left( \bm{x} \right) \right)$, which is expected to hold the style of $\mathcal{Y}$.

Recent methods have achieved higher quality by disentangling the content and style representations \cite{DBLP:conf/eccv/LeeTHSY18,DBLP:conf/eccv/HuangLBK18}.
For example,  Lee et al.\ \shortcite{DBLP:conf/eccv/LeeTHSY18} prepared three latent spaces: $\mathcal{Z}^{c}$ for the content representation, and $\mathcal{Z}_{\mathcal{X}}^{s}$ and $\mathcal{Z}_{\mathcal{Y}}^{s}$ for the style representations of $\mathcal{X}$ and $\mathcal{Y}$, respectively.
Using a content encoder $E_{\mathcal{X}}^{c}\!: \! \mathcal{X} \! \rightarrow \! \mathcal{Z}^{c}$, a style encoder $E_{\mathcal{Y}}^{s}\!: \! \mathcal{Y} \! \rightarrow \! \mathcal{Z}_{\mathcal{Y}}^{s}$, and a generator $G_{\mathcal{Y}}\!: \! \mathcal{Z}^{c} \! \times \! \mathcal{Z}_{\mathcal{Y}}^{s} \! \rightarrow \! \mathcal{Y}$, we can obtain the transferred result $\bm{y} = G_{\mathcal{Y}} ( E_{\mathcal{X}}^{c} \! \left( \bm{x} \right), E_{\mathcal{Y}}^{s} \! \left( \hat{\bm{y}} \right) )$ that resembles the style of a specific reference sample $\hat{\bm{y}} \in \mathcal{Y}$.
As described in \secref{proposed-framework}, our framework basically assumes the availability of the style encoder $E_{\mathcal{Y}}^{s}$ for defining a perceptual metric.
Note that we do not use the generator since provided parametric transformations take its role in our framework.

Though these methods succeeded in yielding high-fidelity results, when it comes to their use for design support, they would not be sufficient in terms of explorability.
One possible approach to allow the exploration of the variations of the result is blending the latent representation of multiple reference samples \cite{DBLP:conf/eccv/XiaoHM18}.
However, exploring a blend parameter between latent representations until finding a satisfactory result can be counterintuitive for users, especially considering the nonlinearity of the latent space \cite{DBLP:journals/corr/abs-1805-07632}.
Another approach is modifying the architecture of GANs to allow a user to specify attributes in the style to be transferred \cite{DBLP:journals/tip/HeZKSC19,DBLP:journals/corr/abs-1912-12396}.
However, since these methods are designed to explicitly learn the attribute information, training data must be annotated with all attributes that can be transferred, which would increase the preparation cost.
Moreover, this requirement implies that the set of transferable attributes is limited by the annotation and not customizable by users.

Therefore, to leverage recent style transfer techniques in exploratory design processes, we propose a new framework to transcribe a transfer effect into parametric transformations available in a tool a user is familiar with.
We believe that our framework connects the recent advances in deep learning and computational design support from a human-centric view.

\section{Parametric Transcription}
\label{sec:proposed}

Our goal is to provide computational support for exploratory design processes by leveraging deep learning techniques.
Before introducing our framework, we first discuss requirements and how parametric transcription satisfies them.

\subsection{Requirements}
\label{sec:proposed-requirements}

It is important to understand that design is often performed exploratorily \cite{DBLP:journals/tog/TaltonGYHK09}.
Let us get back to the example of photo editing in \secref{introduction}.
In this case, the user has two photos, i.e., original and reference photos, and wants to exploratorily edit the original photo by using the reference photo as a loosely specified initial goal.
The user may first try to make some attributes in the style of the original photo (e.g., brightness or contrast) look like those of the reference photo, but then the user may want to change the attributes to modify or to change the degree of the imitation randomly.
Through iterations of such unstructured trials and errors, the user refines their understanding of the space of possible designs around the original and reference photo and progressively establishes the precise form of the design goal.

However, as discussed in \secref{background}, the current deep learning techniques for style transfer are not designed to serve such exploratory design processes.
In particular, the end-to-end transfer methods would not allow exchangeable combinations of step-by-step explorations consisting of trials and errors.
Moreover, they would not facilitate a user's intuitive understanding of the possible design space, as how such a transfer affects the result is not easily predictable by the user.

These points conversely reveal the requirements in style transfer to enable exploratory design processes as follows.
\begin{description}[leftmargin=0.8em,topsep=3pt,itemsep=0pt]
    \item[Parametric] The entire process should be parametrized into a set of manipulable parameters.
    \item[Transparent] The parameters should be easily understandable, allowing users to predict how each parameter affects the result, sometimes even without actually trying them, and grasp what the entire design space looks like.
    \item[Non-destructive] The parameters should be able to be applied in any order to encourage users to perform unstructured trials and errors, such as revoking a specific effect that was manipulated at the beginning of the design process. 
\end{description}
To satisfy them, we propose \textit{parametric transcription}, in which an end-to-end style transfer effect from a reference sample to an original sample is transcribed into parametric transformations available in a tool that a user uses daily.
The user can approximate the style transfer effect in the tool by applying the transcribed parameters to the corresponding transformations and continue further edits such as disabling some transformations that change the attributes the user wants to preserve and tuning the parameters to explore variations.
The key point is to use parametric transformations in existing content editing tools as an interactive generator in style transfer, which meets the above three requirements in the most desirable way from the users' viewpoint.

These requirements also imply that this approach is incomparable with conventional style transfer techniques.
We want to emphasize that, regardless of how such deep learning techniques improve the transfer quality, the reference sample is just the first, loosely specified, goal in the exploratory design processes, and thus, it is crucial to enable exploration to allow users to refine their goal iteratively.
In addition, even if deep learning techniques could provide pretty intuitive controls on the result, it is unlikely to be more intuitive than using parametric transformations in a familiar tool, which are carefully formulated by the tool's developers with domain knowledge.

\subsection{Proposed Framework}
\label{sec:proposed-framework}

We consider an original input $\bm{x}$ and a reference sample $\hat{\bm{y}}$ as well as a set of parametric transformations $\{ T_1 \left( \cdot; \theta_1 \right), \allowbreak T_2 \left( \cdot; \theta_2 \right), \allowbreak \cdots, \allowbreak T_N \left( \cdot; \theta_N \right) \}$ that are controlled by parameters $\bm{\theta} = \{ \theta_1, \allowbreak \theta_2, \allowbreak \cdots, \allowbreak \theta_N \}$.
$N$ denotes the number of transformations available in an existing tool to be used for parametric transcription.
We want to optimize $\bm{\theta}$ so as to make the transformed result $\bm{y} = T_1 \circ T_2 \circ \cdots \circ T_N \left( \bm{x}; \bm{\theta} \right)$ as similar to $\hat{\bm{y}}$ in terms of their styles as possible.\footnote{For simplicity, the composed transformation $T_1 \circ T_2 \circ \cdots \circ T_N \left( \cdot; \bm{\theta} \right)$ is regarded to behave as $\mathcal{X} \rightarrow \mathcal{Y}$.}

Our framework assumes a pretrained model that can measure the perceptual distance in their styles.
For example, an encoder network $E_{\mathcal{Y}}^{s} \left( \cdot \right)$ that outputs the latent representation of the style, such as the ones mentioned in \secref{background}, can be used here.
In other words, what we have been calling \textit{style} is defined by the pretrained model, or indeed, the data used for its training.
Now, the optimization target can be formulated as
\begin{equation*}
\addtolength\abovedisplayskip{-0.6ex}
\addtolength\belowdisplayskip{-0.6ex}
\hat{\bm{\theta}} = \mathrm{argmin}_{\bm{\theta}} \| E_{\mathcal{Y}}^{s} \left( T_1 \circ T_2 \circ \cdots \circ T_N \left( \bm{x}; \bm{\theta} \right) \right) - E_{\mathcal{Y}}^{s} \left( \hat{\bm{y}} \right) \| .
\end{equation*}
In this case, since the encoder network in GANs is trained to output latent variables that match a specific prior distribution, we can choose an appropriate norm function to measure their distance depending on it.
For example, $L_1$-norm is empirically known to perform well for the comparison of latent variables from the prior $\mathcal{N}(0, 1)$ \cite{DBLP:conf/eccv/HuangLBK18}.

In sum, our framework is characterized by four elements: original input, reference sample, set of parametric transformations, and perceptual metric.
The goal of this framework is to exploit the provided transformations applied to the original input so as to obtain a transformed result having a style similar to that of the reference sample.
This would be easy for humans but is challenging for computers because the transformed result cannot be directly compared to the reference sample, as they have different contents.
We thus introduce the perceptual metric leveraging a pretrained model, which enables the comparison between their styles and the optimization of the parameters of the transformations as formulated above.
By exchanging the four elements, our framework can be applied to various situations, as we exhibit from \secref{instagram}--\ref{sec:blender}.

Another key point is the use of black-box optimization algorithms, which frees the proposed framework from constraint on the properties of parametric transformations, such as their differentiability or smoothness.
At the same time, it enables users to freely replace the set of parametric transformations so that they can utilize various existing tools to meet their demand.
Here, we employ \emph{Bayesian optimization} \cite{DBLP:journals/pieee/ShahriariSWAF16} since it is efficient in terms of the number of function evaluations and allows discrete variables to be optimization targets.
In the following scenarios, we used Optuna \cite{DBLP:conf/kdd/AkibaSYOK19} owing to its ease of integration.

Note that the pretrained model used for the perceptual metric is not limited to the encoder network in GANs but can be based on other networks, such as variational autoencoders \cite{DBLP:journals/corr/Doersch16} and flow-based models \cite{DBLP:conf/nips/KingmaD18}, as long as they map the style to a latent representation well.
We emphasize that it is also an advantage of our framework that we do not have to design and train such models if appropriate ones are publicly available.
In fact, for the following scenarios, we have used public models available on GitHub and never trained for this aim.
Moreover, this independence from specific models allows us to incorporate with unseen networks to be proposed in the near future for domains other than those the following scenarios consider.

\section{Experiment: Parametric Transcription of Photo Style Transfer in Instagram}
\label{sec:instagram}

We first present the scenario of transcribing photo style transfer to examine the feasibility of our framework.
To demonstrate that it can employ various tools as a source of parametric transformations, we consider transcribing into Instagram.

\subsection{Related Work}
\label{sec:instagram-related}

There are many methods that leverage deep learning techniques for photo style transfer \cite{8732370}.
For example, Li et al.\ \shortcite{DBLP:conf/cvpr/LiLK019} extended the method proposed by Luan et al.\ \shortcite{DBLP:conf/cvpr/LuanPSB17} to incorporate with an evaluation network, in an analogous manner to GAN-based methods described in \secref{background}.
As we have discussed there, however, these end-to-end methods are not designed for facilitating the intuitive exploration of the variations of the transferred results.
We acknowledge that there are some studies that predict suitable parameters, such as saturation and brightness, for photo enhancement instead of performing such an end-to-end transfer \cite{DBLP:conf/cvpr/OmiyaHSI019,DBLP:journals/tog/HuHXWL18}.
Nevertheless, these methods are aimed at obtaining parameters to realize a professional-like appearance by assuming the availability of a dataset of professionally edited photos.
They thus are not able to serve exploratory processes that use a reference photo as a loosely specified goal.

\subsection{Implementation}
\label{sec:instagram-implementation}

In this scenario, we consider optimizing the parameters of transformations available in Instagram.
However, Instagram provides its photo editing functions in iOS and Android apps and does not provide APIs.
Thus we used UIAutomator, a testing library for Android apps, to control the Instagram app running on an Android emulator.
For a perceptual metric, we used a pretrained model (encoder network) of neural photo style transfer published by Li et al.\ \shortcite{DBLP:conf/cvpr/LiLK019}, which is trained to encode a reference photo into its style representation during the end-to-end photo style transfer.

In each optimization step, our framework first samples parameters to try next from Optuna \cite{DBLP:conf/kdd/AkibaSYOK19}.
Then, via UIAutomator, it applies all transformations, such as adjusting brightness or vignette, as well as choosing a photo filter from 24 provided filters based on the sampled parameters.
The transformed photo is cropped from the screenshot and provided to the pretrained model to compare its style with that of the reference photo.
Finally, the perceptual distance to the reference photo is returned to Optuna as a score to be minimized until 1,000 iterations are performed.
Note that Instagram yields the same result for the same set of parameters without regard to the order of manipulating the parameters on the interface, which freed us from optimizing the order.

\subsection{Subjective Evaluation}
\label{sec:instagram-evaluation}

We evaluated the imitating quality of our framework in comparison with that of humans.
For the same pair of original and reference photo we obtained three imitating photos, one by the proposed framework and the others by human participants.
The imitating photos were evaluated in regard to their similarity to the reference photo by human evaluators recruited using Amazon Mechanical Turk (AMT).

\subsubsection{Procedure}

\begin{figure}[t]
    \centering
    \includegraphics[width=0.9\columnwidth]{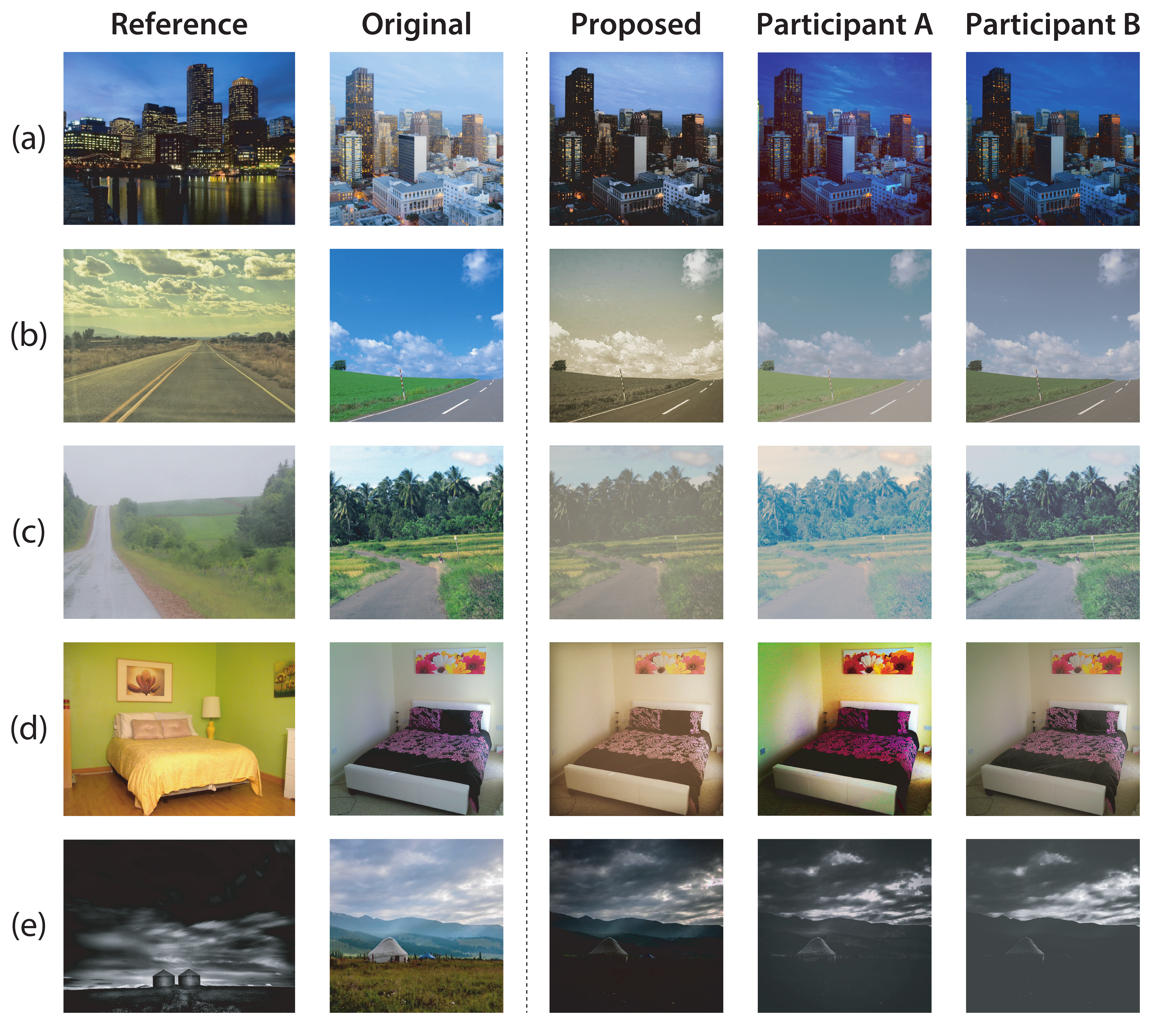}
    \caption{Examples of the original and reference photos and the imitating photos obtained by our framework and the participants.}
    \label{fig:instagram}
\end{figure}

We first prepared imitating photos using our framework for ten pairs of original and reference photos, which were taken from samples released by Luan et al.\ \shortcite{DBLP:conf/cvpr/LuanPSB17}.
For comparison baselines, we recruited two participants who had experience in using Instagram and asked them to edit the same original photo, using Instagram, to look like the corresponding reference photo.
\figref{instagram} shows some examples.
Note that comparison with the previous methods for photo style transfer would not be meaningful because, as described in \secref{proposed-requirements}, they perform an end-to-end transfer whereas our aim is finding parameters for transformations available in Instagram to enable exploratory design processes.

Then we recruited 30 evaluators in AMT and asked them to fill a questionnaire consisting of ten questions.
In each question, they were presented with a reference photo and three imitating photos, one of which was obtained by the proposed framework and the others by the human participants, and were instructed to rank them in order of their similarity to the reference photo.
Their responses were analyzed using aligned ranked transform (ART) \cite{doi:10.1080/03610919308813085}, which can analyze ranked data nonparametrically.
This allows us to examined the effect of the ways to obtain the imitating photos (i.e., by our framework or by the human participants) in a manner similar to a two-way ANOVA test.

\subsubsection{Results}

\begin{table*}[t]
    \caption{Distribution of the collected responses indicating how many times the imitating results in each group (obtained by the proposed framework or by the human participants) were placed at each rank.}
    \label{tab:result}
    \renewcommand{\arraystretch}{0.95}
    \centering
    \footnotesize
    \begin{tabular}{lccc}
        \toprule
        \textbf{(a)} \secref{instagram} & 1st & 2nd & 3rd \\
        \midrule
        Proposed                        & 105 &  87 & 108 \\
        Participant A                   &  92 &  99 & 109 \\
        Participant B                   & 103 & 114 &  83 \\
        \bottomrule
    \end{tabular}
    \hspace{1em}
    \begin{tabular}{lccc}
        \toprule
        \textbf{(b)} \secref{snow}      & 1st & 2nd & 3rd \\
        \midrule
        Proposed                        & 118 &  92 &  90 \\
        Participant A                   & 113 & 113 &  74 \\
        Participant B                   &  69 &  95 & 136 \\
        \bottomrule
    \end{tabular}
    \hspace{1em}
    \begin{tabular}{lccc}
        \toprule
        \textbf{(c)} \secref{blender}   & 1st & 2nd & 3rd \\
        \midrule
        Proposed                        &  93 &  89 & 118 \\
        Participant A                   &  98 & 104 &  98 \\
        Participant B                   & 109 & 107 &  84 \\
        \bottomrule
    \end{tabular}
    \vspace{-6pt}
\end{table*}

The distribution of the collected responses is shown in \tabref{result}(a).
The ART analysis indicated no main effect from how the imitating photos were obtained, and thus implied that the proposed framework demonstrated performance comparable to that of humans in imitating the style of the reference photo.
In other words, our framework is feasible for the scenario of transcribing photo style transfer.

We also examined agreement among the evaluators for each question by using Kendall's $W$.
As a result, \figref{instagram}(d) and (e) were two of the cases that showed the lowest agreement ($W = 0.021$ and $0.048$, respectively).
Looking at the imitating photos for these cases, the dispersion of the responses for \figref{instagram}(e) can be attributable to the almost identical appearance.
On the other hand, the imitating photos in \figref{instagram}(d) evoked the dispersion though they seemed to vary in appearance.
We suppose that this was caused by the discrepancy between the original and reference photos, which made it difficult to imitate the style using Instagram, and thus none of the imitating photos was similar enough to obtain the consensus of the evaluators.
This dispersion would be resolved by using editing tools with more expressiveness than Instagram.

\section{Experiment: Parametric Transcription of Facial Makeup Transfer in SNOW}
\label{sec:snow}

We next consider transcribing facial makeup transfer into SNOW using the proposed framework.
SNOW is one of the popular iOS and Android apps providing functions for editing selfies with various makeup features.

\subsection{Related Work}
\label{sec:snow-related}

As with photo style transfer, facial makeup transfer has involved deep learning techniques, such as GANs \cite{DBLP:conf/iccv/GuWCTT19} or flow-based models \cite{DBLP:conf/cvpr/ChenHWTSC19}.
For example, Gu et al.\ \shortcite{DBLP:conf/iccv/GuWCTT19} exploited disentangled latent representations of a face content and a makeup style, as we described in \secref{background}.
However, these methods aimed to obtain a transferred selfie but not to allow users to explore the variations of the results, which motivated us to apply parametric transcription.

\subsection{Implementation}
\label{sec:snow-implementation}

Since SNOW also does not provide APIs, we used the Android emulator with UIAutomator, as in \secref{instagram-implementation}.
For a perceptual metric, we used a pretrained model (encoder network) of neural facial makeup transfer published by Gu et al.\ \shortcite{DBLP:conf/iccv/GuWCTT19}, which is trained to encode a reference selfie with makeup into the latent representation of its makeup style.
Here, although the original end-to-end facial makeup transfer had a generator to generate a facial image from this latent representation, we did not use it and used only the encoder.
Then we used Optuna for 1,000 iterations to find parameters of the makeup transformations, such as lip color and eyebrows, that minimize the distance to the reference selfie in the latent space.

\subsection{Subjective Evaluation}
\label{sec:snow-evaluation}

\subsubsection{Procedure}

\begin{figure}[t]
    \centering
    \includegraphics[width=0.9\columnwidth]{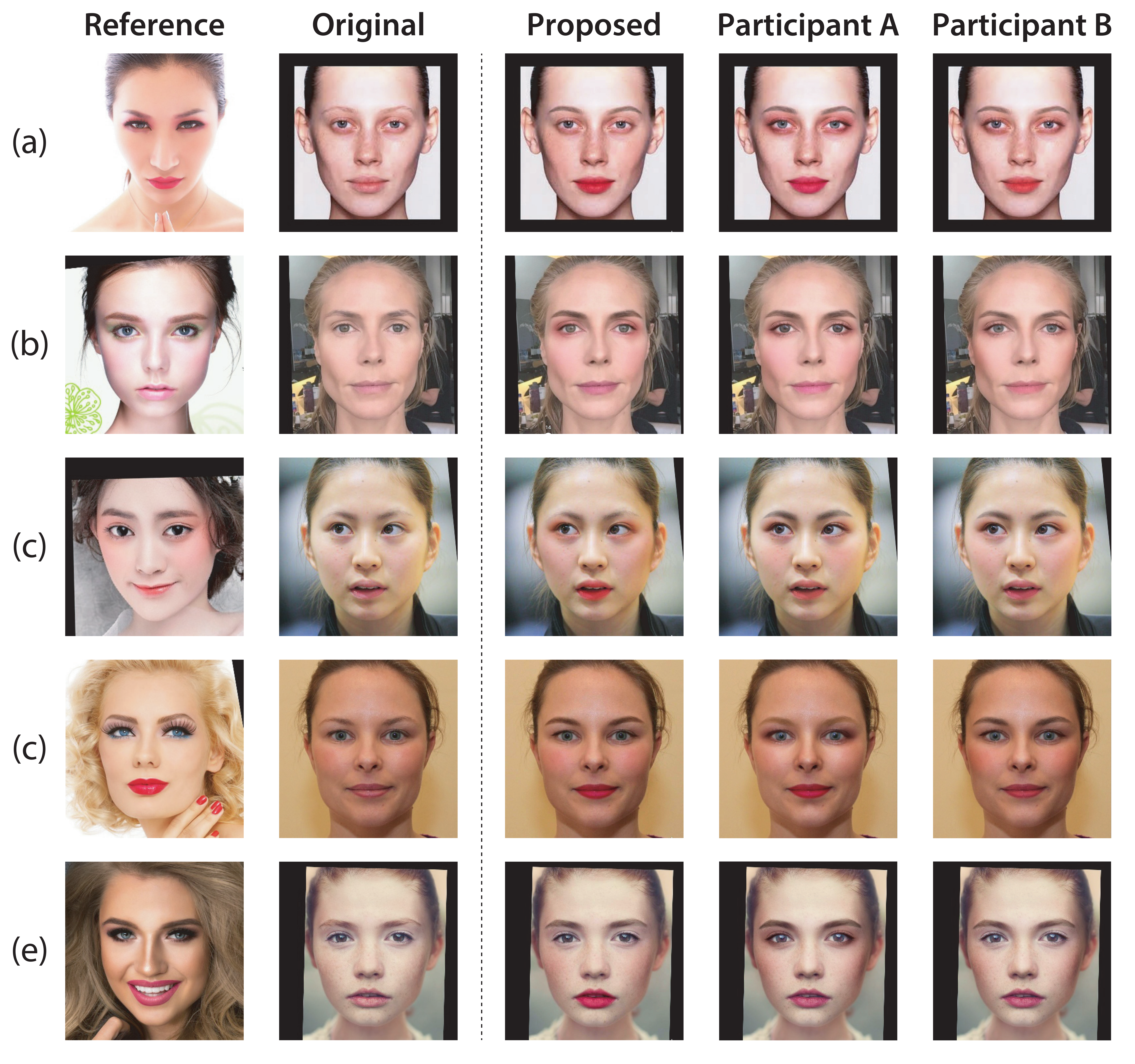}
    \caption{Examples of the original and reference selfies and the imitating selfies obtained by our framework and the participants.}
    \label{fig:snow}
\end{figure}

We conducted a subjective evaluation in the same manner as described in \secref{instagram-evaluation}.
We randomly selected ten pairs of the original and reference selfies from the dataset of Gu et al.\ \shortcite{DBLP:conf/iccv/GuWCTT19} and prepared imitating selfies by using our framework.
We also recruited two participants and asked them to edit the original selfies to look like the reference selfies by using SNOW.
\figref{snow} shows some examples.
Then we again recruited 30 evaluators in AMT and asked them to rank the imitating selfies ten times in the same manner as before.
Their responses were later analyzed using ART.

\subsubsection{Results}

The distribution of the collected responses is shown in \tabref{result}(b).
Different from \secref{instagram-evaluation}, the main effect of how the imitating selfies were obtained was observed ($p \! < \! 0.001$).
Thus we conducted Tukey's post-hoc analysis and found significant differences between the selfies edited by Participant B and those either edited by Participant A or obtained by the proposed framework ($p \! < \! 0.001$) but no significant difference between selfies edited by Participant A and those obtained by the proposed framework ($p \! = \! 0.57$).
Therefore, given \tabref{result}(b), the proposed framework outperformed one human participant and showed performance comparable to that of the other in regard to imitating the makeup style of the reference selfie.
Along with \secref{instagram-evaluation}, this result suggests that our framework is feasible in various situations and can automatically obtain parameters yielding human-level quality.
In other words, even without any additional data or training, it can enable users to leverage style transfer techniques exploratorily within existing tools by presenting the parameters.

\section{Experiment: Parametric Transcription of Text-to-Object Composition in Blender}
\label{sec:blender}

\begin{figure}[t]
    \centering
    \includegraphics[width=0.98\columnwidth]{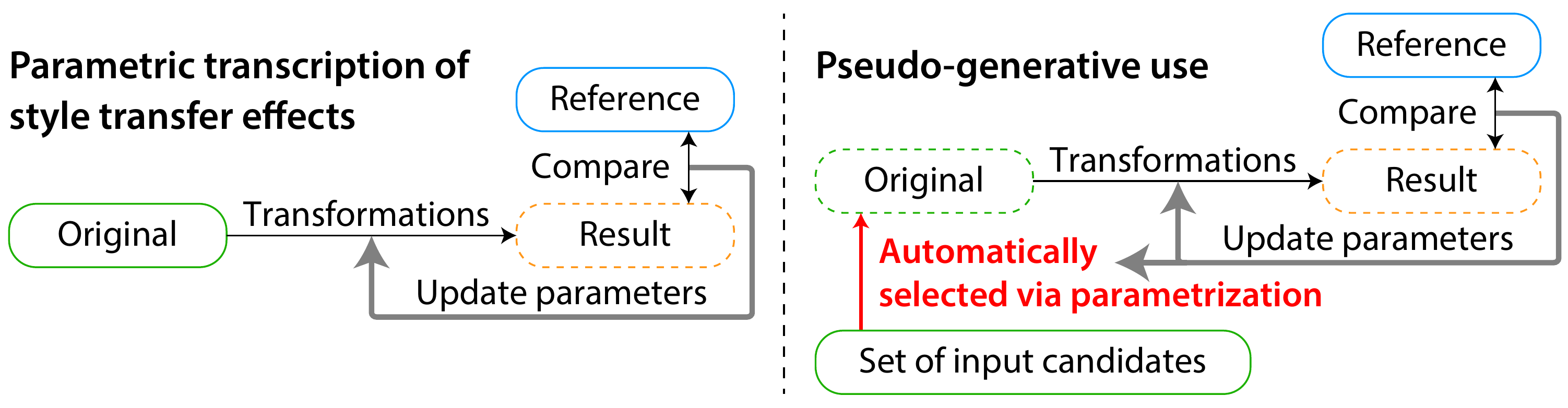}
    \caption{Pipeline of applying the proposed framework for a pseudo-generative purpose. It optimizes not only the parameters of the transformations but also the selection of the original input.}
    \label{fig:generative}
\end{figure}

The application domain of our framework is not limited to images but includes 3D objects.
In addition, assuming a set of input candidates, we can also use the proposed framework for a pseudo-generative purpose, in which the original input is selected from the set.
As shown in \figref{generative}, this selection process is parameterized and directly integrated into the optimization of the parameters of the transformations.

We here examine a scenario of composing a 3D object as specified in a sentence.
That is, given an instruction sentence and a set of candidate 3D objects, our framework selects an appropriate one and optimizes the materials of the object.
The fidelity of the composed object to the instruction sentence is measured by using a pretrained model for text-to-image synthesis as a perceptual metric because, to the best of our knowledge, there is no established method for text-to-object composition.
We emphasize that this scenario illustrates the flexibility of our framework, which can deal with situations beyond the original task that the pretrained model aims at.

\subsection{Related Work}
\label{sec:blender-related}

Although there are many methods for text-to-image synthesis \cite{DBLP:conf/icml/ReedAYLSL16,DBLP:conf/iccv/ZhangXL17}, few have tried generating 3D objects from a sentence.
For example, Chen et al.\ \shortcite{DBLP:conf/accv/ChenCSCFS18} and Fukamizu et al.\ \shortcite{DBLP:journals/corr/abs-1901-07165} leveraged GANs to generate 3D voxels from a sentence.
However, since these methods produce voxelized objects, which do not hold a high-level structure, they would be less suitable for explorative design processes than other structured formats would.
On the other hand, though our framework requires a set of candidate objects, it ensures the explorability of the variations of its output by preserving the original structure of the objects, such as polygon meshes.
Even though there is no existing method for text-to-editable 3D object composition, we enabled this composition by leveraging an existing pretrained model for text-to-image synthesis to measure the fidelity.

\subsection{Implementation}
\label{sec:blender-implementation}

As this scenario is not about style transfer, we extended the definition of the perceptual metric.
Here the perceptual metric is required to measure the deviation from how the resulting 3D object should look as specified in the instruction sentence.
For this purpose, we used the discriminator (not encoder) of StackGAN \cite{DBLP:conf/iccv/ZhangXL17}, a text-to-image synthesis model.
Like other GAN-based techniques mentioned in \secref{background}, StackGAN consists of a generator to generate an image and a discriminator to judge whether a given image is a real or generated one from the viewpoint of a given sentence.
Interestingly, we could directly leverage the discriminator for the perceptual metric, since it could evaluate how well a given image matches how the image should look like as specified in the sentence.
Thus, once we convert the composed 3D object into an image by rendering, we can measure the perceptual distance by feeding the rendered image to the discriminator.

Specifically, our framework controls Blender via its Python API to render an object from a fixed point of view.
The rendered image is then provided to one of the pretrained models of StackGAN; in particular, we used the one for text-to-bird image synthesis.
The distance score obtained by the discriminator is returned to Optuna, which decides the next parameters for choosing an object and the color of all materials in the object so as to minimize the score, for 1,000 iterations.

\subsection{Subjective Evaluation}
\label{sec:blender-evaluation}

\subsubsection{Procedure}

\begin{figure}[t]
    \centering
    \includegraphics[width=\columnwidth]{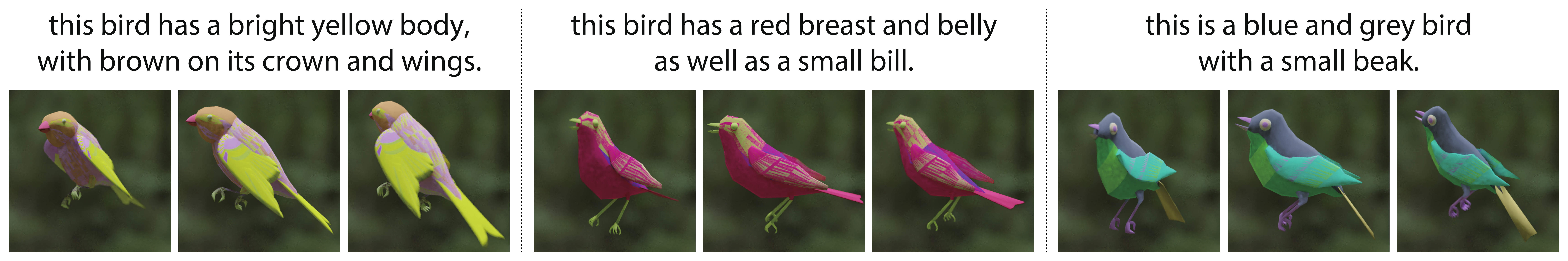}
    \caption{Detailed examples of the instruction sentences and the composed 3D objects obtained by using the proposed framework.}
    \label{fig:blender-example}
\end{figure}

\begin{figure}[t]
    \centering
    \includegraphics[width=0.95\columnwidth]{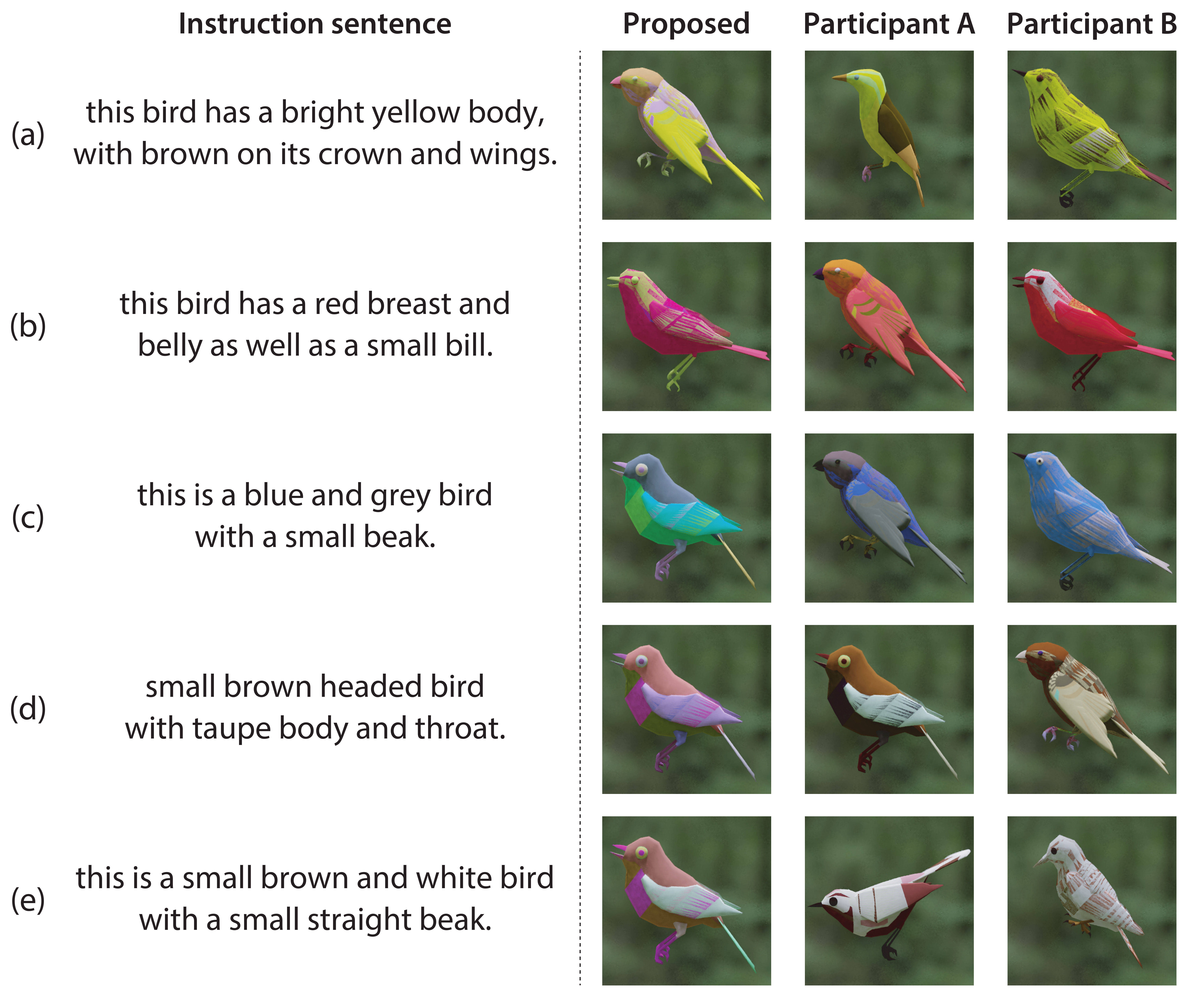}
    \caption{Examples of the instruction sentences and the 3D objects obtained by our framework and the participants.}
    \label{fig:blender}
\end{figure}

We first prepared ten instruction sentences and obtained a 3D object for each using our framework.
For the instruction sentences, we randomly selected ones describing the appearance of birds from the same dataset \cite{DBLP:conf/icml/ReedAYLSL16} that Zhang et al.\ \shortcite{DBLP:conf/iccv/ZhangXL17} used.
For the candidate objects, we prepared nine 3D objects categorized as birds in the ShapeNetSem dataset \cite{DBLP:conf/cvpr/SavvaCH15}.
Each of them contained 6 to 12 materials whose diffuse colors were optimized via the three RGB parameters to make the object look as specified in the instruction sentence.
Some examples of the instruction sentences and the composed objects are presented in \figref{blender-example}.
We then conducted the subjective evaluation in the same manner as in Sec.~\ref{sec:instagram-evaluation} and \ref{sec:snow-evaluation}.
We asked two participants who had experience in using Blender to compose a 3D object so as to match with the provided sentence by choosing one from the same set of candidate objects and adjusting the colors of its materials (\figref{blender}).
We also recruited 30 evaluators in AMT and asked them to rank the rendered images of the composed 3D objects in order of their fidelity to the instruction sentences.

\subsubsection{Results}

The distribution of the collected responses is shown in \tabref{result}(c).
As in \secref{snow-evaluation}, the main effect of how the composed objects were obtained was observed ($p \! = \! 0.017$).
Thus we conducted Tukey's post-hoc analysis and found significant differences between the objects composed by Participant B and those obtained by the proposed framework ($p \! = \! 0.014$) but no significant difference between the objects composed by Participant A and those either composed by Participant B or obtained by the proposed framework ($p \! = \! 0.62$ and $0.14$).
In other words, whereas the proposed framework was outperformed by one of the human participants regarding the fidelity to the instruction sentences, its performance was still comparable to that of the other participant.
We thus conclude that, while the proposed framework may not yet achieve the human-level quality in this pseudo-generative use, its quality can be reasonably close to the human level and thus enough for a starting point of exploratory design processes.

\section{Discussion}
\label{sec:discussion}

We have demonstrated the feasibility of the proposed framework in various scenarios.
To facilitate its further application, we discuss how we can leverage it, especially for computational design support.
We first situate our framework by referring to related work and then discuss possible application scenarios of parametric transcription as well as its limitations.

\subsection{Existing Optimization-Based Techniques for Computational Design Support}
\label{sec:discussion-optimization}

Prior to our framework, many optimization-based techniques have been proposed for computational design support \cite{DBLP:conf/siggraph/MarksABFGHKMPRRSS97,DBLP:journals/tvcg/SedlmairHBPM14} given that most design processes are parameterized by a set of either continuous or discrete parameters in current editing tools (e.g., using control panels including many sliders) \cite{Koyama2018}.
In particular, researchers have developed human-in-the-loop optimization techniques to deal with fuzzy optimization criteria that involve perceptual metrics or preference.

For example, evolutionary computation has been leveraged in combination with an iterative human evaluation to solve perceptual optimization problems \cite{949485}.
More recently, Bayesian optimization has been used for human-in-the-loop optimization frameworks \cite{DBLP:conf/nips/BrochuFG07,DBLP:journals/tog/KoyamaSG20}.
This is because it requires only a small number of function evaluations to find a good solution \cite{DBLP:journals/pieee/ShahriariSWAF16} and thus reduces the amount of human evaluation required when used in human-in-the-loop optimization.

For situations where the distance to the reference sample can be calculated by computers, optimization-based methods that work without human evaluation have been employed.
Such an approach for fitting parameters to replicate a reference sample via optimization is often referred to as \emph{inverse design} \cite{DBLP:conf/siggraph/MarksABFGHKMPRRSS97}.
Moreover, inverse \emph{procedural} modeling \cite{DBLP:journals/tog/TaltonLLDMK11,DBLP:journals/tog/HuDR19} searches for an optimal combination of parametric procedures and their parameters for replicating a given reference sample.
One of its primary motivations is to support further editing of the result by changing the parameters, which is common to our motivation.

These frameworks differ from ours, however, in that ours does not assume that the reference sample and replicating result can be directly compared (e.g., by pixel-wise comparison).
In contrast, we have considered transcribing style transfer effects, in which even applying the optimal transformations to the original input does not yield a transformed result that is identical to the reference sample because the result and the sample have different contents.
Thus, while inverse design tries to exactly reproduce the reference sample based on a direct comparison, our framework is required to perform the optimization based on a content-invariant comparison.
We addressed this point without the help of human evaluation by leveraging a pretrained model to obtain a perceptual metric, which enabled new application scenarios that are intractable to conventional inverse design approaches.

Moreover, the tool- and domain-agnostic applicability of our framework is another of its advantages.
In particular, we acknowledge a recent method in inverse design that involves a perceptual metric \cite{DBLP:journals/tog/ShiLHSBMM20}.
It optimizes a procedural texture model to resemble a reference texture in a similar manner as Hu et al.\ \shortcite{DBLP:journals/tog/HuDR19}.
On the other hand, this method leverages a perceptual metric extracted from a pretrained model to perform a style-aware comparison instead of the pixel-wise comparison and in this sense is analogous to our framework.
Still, this method is specifically designed for a custom-made texture model, as it relies on the differentiability of the texture model.
In contrast, our framework is not only tool- and domain-agnostic but also potentially suitable for networks to be proposed in the near future.

\subsection{Application for Computational Design Support}
\label{sec:discussion-design}

As we discussed in \secref{proposed-requirements}, our framework was designed to enable exploratory design processes.
In this regard, Shneiderman\ \shortcite{DBLP:journals/cacm/Shneiderman07a} pointed out that supporting exploratory processes is one of the principal functions required for creativity support tools accelerating discovery and innovation.
In fact, many creativity support tools enabling exploratory design in various situations have been proposed \cite{DBLP:conf/uist/HartmannYAYK08,DBLP:journals/tog/TaltonGYHK09}.
Our approach can complement this research direction in terms of how to leverage the end-to-end style transfer techniques for creativity support tools.

The essence of the proposed approach is to enable explorations within a tool the user uses daily by imitating a reference sample with the help of the deep learning techniques, rather than directly using the techniques.
It would allow not only the seamless connection with familiar design processes but also a new design process starting from the imitation.
For example, since the proposed framework provides not only the imitating result but also the parameters of the transformations to imitate, the user can utilize them in various ways other than making the original sample similar to the reference sample, e.g., making it dissimilar by applying the obtained parameters inversely.
Such imitation-driven explorations would provide new inspirations for creators, as Dal{\'{i}}\ \shortcite{Dali1970} emphasized the importance of imitation as a source of creativity when he said that ``those who do not want to imitate anything, produce nothing.''
From these points, though further investigations are demanded, we believe that parametric transcription has great potential to establish a new form of creativity support tools.

Furthermore, our framework can be exploited for computational design support in a way analogous to that in which sound sampling techniques have been used in musical scenes.
That is, the user can store the transcribed parameters of style transfer effects and use them at other times as if they were a custom filter.
In fact, such a sampler-based approach for computational design support has been studied by many researchers; for example, I/O Brush \cite{DBLP:conf/chi/RyokaiMI04} enables a user to pick up the surface of any object and use it as a brush in digital drawing.
Our framework can be more flexible since it performs not just sampling but transcribing parametrically, which allows the user to edit the sampled parameters.

\subsection{Limitations}
\label{sec:discussion-limitations}

Our framework has some limitations.
First, the running time can be a hurdle for its practical use.
For example, in the case of transcribing photo style transfer into Instagram, it took about an hour for 1,000 iterations.
This is partially due to the black-box optimization but mainly due to the overhead caused in manipulating the Instagram app on an Android emulator using the testing library.
As in humans editing photos in Instagram, it takes at least a second for a single iteration applying all parameters on the interface of the app.
The running time can be shortened by performing the optimization process in parallel or by targeting applications that provide APIs (e.g., Photoshop or Blender) to reduce the overhead.

In addition, the quality of parametric transcription is constrained by the expressiveness of available parametric transformations.
When the styles of the original input and the reference sample are so different that they are difficult to imitate with any combination of the available transformations, the imitating results will not be so faithful.
In particular, content-aware style transfer, such as artistic portrait stylization \cite{DBLP:journals/tog/YanivNS19}, would be challenging to be imitated by parametric transformations available in existing photo editing tools.

Our framework does not cover domains for which there is no tool providing appropriate parametric transformations.
For instance, there are some studies on transferring the style of human motions \cite{DBLP:journals/tog/Mitra16,DBLP:journals/cga/HoldenHKK17}.
Still, it is difficult to apply our framework to transcribing motion style transfer parametrically because such human motion data is usually represented by the joint positions on a frame-by-frame basis.
Thus, there is no established mechanism to control it with a set of parametric transformations.

Nevertheless, the exploration of further applications is a fruitful endeavor.
Evaluating the user experience of designing with the support of parametric transcription would provide implications in promoting human-AI collaboration.
Investigating how users can utilize transcribed results in their design process would also be interesting, as discussed in \secref{discussion-design}.

\section{Conclusion}
\label{sec:conclusion}

This paper introduced a new approach of human-AI collaboration for computational design support---parametric transcription of style transfer effects---and a framework enabling it.
Given an original input and a set of parametric transformations, our framework automatically optimizes their parameters to make the transformed result have a style similar to that of a reference sample by leveraging a perceptual metric retrieved from an existing pretrained model.
Unlike end-to-end style transfers, this framework offers explorability of the variations of the result by providing parameters of the transformations so that users can further edit the result intuitively in a third-party tool.
Through three experiments, we showed that our framework can be applied to various situations, including pseudo-generative cases, while achieving human-comparable performance.
These results teach us that, even without any additional training or data, a simple combination of existing techniques can expand the applicability of numerous pretrained models in a tool- and domain-agnostic manner.

\section*{Acknowledgments}
This work was supported in part by JST ACT-X (JPMJAX200R) and CREST (JPMJCR20D4), Japan.

% Bibliography
\bibliographystyle{named}
\bibliography{paper}

\end{document}